\documentclass[12pt]{article}
\usepackage[letterpaper, portrait, margin=1in]{geometry}

\usepackage{microtype}
\usepackage{graphicx}
\usepackage{subfigure}
\usepackage{booktabs} 
\usepackage{amsthm,amsmath,amssymb}
\usepackage{dsfont}
\usepackage{algorithmic,algorithm}

\usepackage{hyperref}
\usepackage{natbib}

\title{Reverse-Engineering Deep ReLU Networks}
\author{David Rolnick \hskip .2in Konrad P.~K\"ording\\
University of Pennsylvania\\
\small{\texttt{\{drolnick,kording\}@seas.upenn.edu}}}
\date{}

\newtheorem{thm}{Theorem}

\newtheorem{lem}{Lemma}

\newcommand{\N}{\mathcal{N}}
\newcommand{\R}{\mathbb{R}}

\newcommand{\Hh}{\mathcal{H}}
\newcommand{\nin}{n_\text{in}}
\newcommand{\nout}{n_\text{out}}
\newcommand{\bb}{{\bf b}}
\newcommand{\xx}{{\bf x}}

\newcommand{\pp}{{\bf p}}
\newcommand{\qq}{{\bf q}}
\newcommand{\vv}{{\bf v}}
\newcommand{\WW}{{\bf W}}
\newcommand{\One}{\mathds{1}}
\DeclareMathOperator{\ReLU}{ReLU}

\begin{document}

\maketitle

\begin{abstract}
It has been widely assumed that a neural network cannot be recovered from its outputs, as the network depends on its parameters in a highly nonlinear way. Here, we prove that in fact it is often possible to identify the architecture, weights, and biases of an unknown deep ReLU network by observing only its output. Every ReLU network defines a piecewise linear function, where the boundaries between linear regions correspond to inputs for which some neuron in the network switches between inactive and active ReLU states. By dissecting the set of region boundaries into components associated with particular neurons, we show both theoretically and empirically that it is possible to recover the weights of neurons and their arrangement within the network, up to isomorphism.
\end{abstract}

\section{Introduction}
A deep neural network computes a function from inputs to outputs, where the structure and parameters of the network control the function that is expressed. While each parameter influences the overall function, this influence is highly nonlinear, and the effects of different neurons within the network would appear superficially to be hopelessly entangled. Consequently, it has been widely supposed in the field that it is impossible to recover the structure and parameters of a network merely by observing its output on different inputs.

Were it possible to reverse-engineer a neural network from the function it computes, there could be serious implications for security and privacy. In many deployed deep learning systems, the output is freely available but the network used to generate that output is not. The ability to recover a confidential network could even expose data used to train the network if such data could be reconstructed from the network's weights.

The related question of reverse-engineering biological neural networks is of foundational interest in neuroscience. Experimental neuroscientists can record the output of some neurons (e.g.~complex cells in primary visual cortex) but not others (e.g.~simple cells) and must also infer the synaptic weights between them. Our understanding of the brain would be greatly improved if it were possible to identify the internal components of a neural circuit based on recordings of the output of that circuit.

In this work, we show mathematically and empirically that it is, in fact, often possible to recover the structure and weights of an unknown ReLU network by querying it. Our approach leverages the fact that a ReLU network is piecewise linear and transitions between linear pieces when one of the ReLUs of the network transitions from its inactive to its active state. For neurons in the first layer of the network, such transitions occur along hyperplanes through input space; the equations of these hyperplanes determine the weights and biases of the first layer (up to sign and constant scaling). For neurons in subsequent layers, transitions occur along ``bent hyperplanes'' that bend where they intersect bent hyperplanes associated with earlier layers. Measuring the intersections between bent hyperplanes allows us to recover the weights between the corresponding neurons.

Our principal contributions are:
\begin{itemize}
    \item We prove that the architecture, weights, and biases of a deep ReLU network can be recovered (up to isomorphism) from the arrangement of regions on which the network is linear.
    \item We describe an algorithm to recover the network by approximating the boundaries between these linear regions, with the only information given about the network being its output on specified queries.
    \item We demonstrate the success of our algorithm in reverse-engineering both trained and untrained ReLU networks.
\end{itemize}

Each of these contributions significantly advances the state of the art. No prior work has, to our knowledge, been able to deduce even the first layer of a fully-connected network with 2 hidden layers. By contrast, our theoretical results and algorithm hold for reverse-engineering any layer of a network of any depth. Furthermore, we show empirically that our algorithm is able to reconstruct the first layer of 2-, 3-, and 4-layer networks, as well as the second layer of 2-layer networks.

\begin{figure*}[ht]
\begin{center}
\includegraphics[scale=0.43]{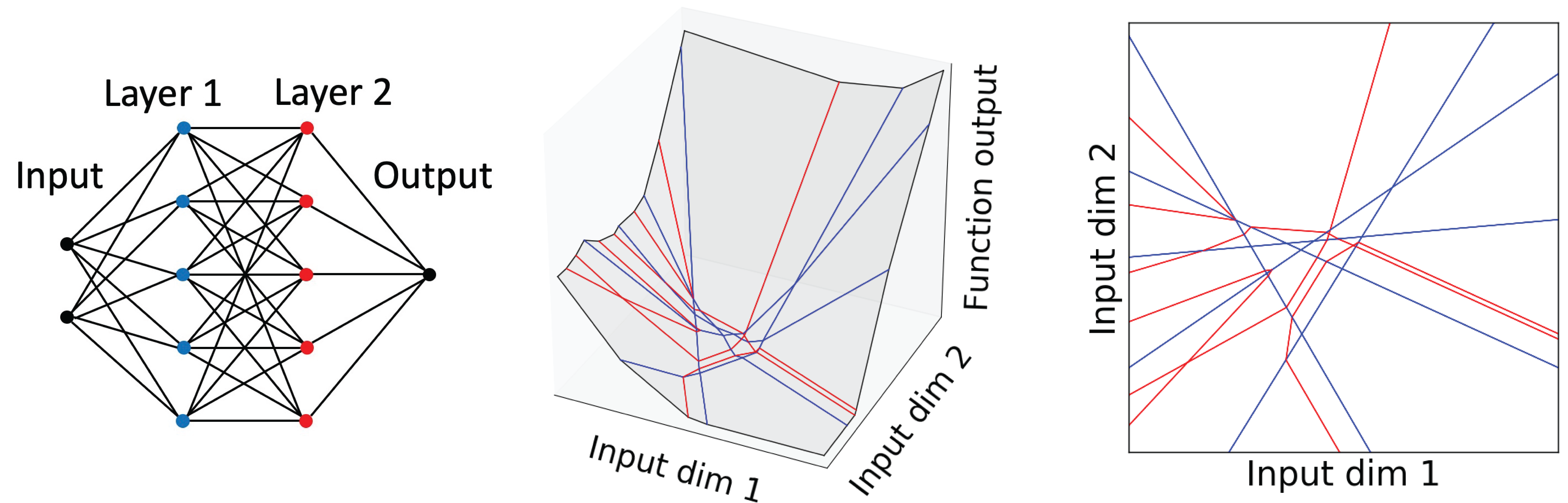}
\caption{Left: Architecture of a ReLU network $\N(\xx):\R^2\to \R$ with two hidden layers of width 5. Center: Graph of the piecewise linear function $\N(\xx)$ as a function of the two input variables. Right: Boundaries between linear regions of $\N$, essentially a ``flattened'' form of the center image. Boundaries $B_z$ corresponding to neurons $z$ from the first and second layers are shown in blue and red, respectively.}
\label{fig:boundaries}
\vspace{-0.2in}
\end{center}
\end{figure*}

\section{Related work}
Various works within the deep learning literature have considered the problem of inferring a network given its output on inputs drawn (non-adaptively) from a given distribution. It is known that this problem is in general hard \citep{goel2017reliably}, though positive results have been found for certain specific choices of distribution in the case that the network has only one or two layers \citep{ge2019learning, goel2017learning}.  By contrast, we consider the problem of reconstructing a network of arbitrary depth, given the ability to issue queries at specified input points. In this work, we leverage the theory of linear regions within a ReLU network, an area studied e.g.~by \citet{telgarsky2015representation, raghu2017expressive, hanin2019complexity}. Most recently, \citet{hanin2019deep} considered the boundaries between linear regions as arrangements of ``bent hyperplanes''. \citet{milli2019model, jagielski2019high} show the effectiveness of this strategy for networks with one hidden layer. For inference of other properties of unknown networks, see e.g.~\citet{oh2019towards}.

Neuroscientists have long considered similar problems with biological neural networks, albeit armed with prior knowledge about network structure. For example, it is believed that complex cells in the primary visual cortex, which are often seen as translation-invariant edge detectors, obtain their invariance through what is effectively a two-layer neural network \citep{kording2004complex}. A first layer is believed to extract edges, while a second layer essentially implements max-pooling. Many biological neurons appear to be well modeled as the ReLU of a linear combination of their inputs \citep{chance2002gain}. \citet{heggelund1981receptive} perform physical experiments akin to our approach of identifying one ReLU at a time, by applying inputs that move individual neurons above their critical threshold one by one. Being able to solve such problems more generically would be useful for a range of neuroscience applications.

\section{Theoretical framework}
\label{sec:prelim}

We will, in general, consider fully connected, feed-forward neural networks (multi-layer perceptrons) with ReLU activations. Each such network $\N$ defines a function $\N(\xx)$ from input space $\R^{\nin}$ to output space $\R^{\nout}$. We denote the layer widths of the network by $\nin$ (input layer), $n_1, n_2,\ldots, n_d$, $\nout$ (output layer). We denote by $\WW^k$ the weight matrix from layer $(k-1)$ to layer $k$, where layer $0$ is the input; and $\bb^k$ denotes the bias vector for layer $k$. Given a neuron $z$ in the network, we use $z(\xx)$ to denote its preactivation for input $\xx \in \R^{\nin}$. Thus, for the $j$th neuron in layer $k$, we have $$z_j^k(\xx) = \sum_{i=1}^{n_{k-1}} \WW^k_{ij}\ReLU(z_i^{k-1}(\xx) + \bb^k_i).$$

\subsection{Isomorphisms of networks}
\label{subsec:isomorphisms}
Before showing how to reverse-engineer the parameters of a neural network, we must consider to what extent these parameters can be inferred unambiguously. For every network, there are a number of other networks that define exactly the same function from input space to output space. We say that two networks $\N$ and $\N'$ (possibly with different parameters or architecture) are \emph{isomorphic} if $\N(\xx)=\N'(\xx)$ for all $\xx\in\R^{\nin}$. For a fully connected network with ReLU activation, there are two obvious network isomorphisms:

\textbf{Permutation.} The order of neurons in each layer of a network $\N$ does not affect the underlying function. Formally, let $p_{k, \sigma}(\N)$ be the network obtained from $\N$ by permuting layer $k$ according to $\sigma$ (along with the corresponding weight vectors and biases). Then, $p_{k, \sigma}(\N)$ is isomorphic to $\N$ for every layer $k$ and permutation $\sigma$.

\textbf{Scaling.} Due to the ReLU's equivariance under multiplication, it is possible to scale the incoming weights and biases of any neuron, while inversely scaling the outgoing weights, leaving the overall function unchanged. Formally, for $z$ the $i$th neuron in layer $k$ and for any $c>0$, let $s_{z,c}(\N)$ be the network obtained from $\N$ by replacing $\WW^k_{\cdot i}$, $\bb^k_i$, and $\WW^{k+1}$ by $c\WW^k_{\cdot i}$, $cb^k_i$, and $(1/c)\WW^{k+1}_{i\cdot}$, respectively. It is simple to prove that $s_{z,c}(\N)$ is isomorphic to $\N$ (see Appendix \ref{app:lem}).

It is worth noting that permutation and scaling are isomorphisms for \emph{every} ReLU network, but some networks may have additional isomorphisms. It is shown in \citet{phuong2019functional} that there exist ReLU networks of every architecture for which all isomorphisms are given by some combination of permutation and scaling.\footnote{See also \citet{fefferman1994reconstructing} for a proof that an analogous set of isomorphisms is comprehensive for networks with tanh activation.} As we demonstrate in \S\ref{subsec:boundaries}, there exists a positive-measure set of ReLU networks with additional isomorphisms.

In this work, we show that, where additional isomorphisms do not occur, it is in fact often possible both in theory and in practice to reverse-engineer ReLU networks up to permutation and scaling. 

\subsection{Linear regions}
\label{subsec:regions}
Consider a network $\N$ and neuron $z\in \N$. We denote by $B_z$ the set of $\xx$ for which $z(\xx)=0$. We call $B_z$ the \emph{boundary} associated with neuron $z$, and we say that $B = \bigcup B_z$ is the \emph{boundary} of the overall network. We refer to the connected components of $\R^{\nin}\setminus B$ as \emph{linear regions}. Thus, each linear region corresponds to a pattern of activations across the ReLUs in the network; within a region, each ReLU computes a fixed linear function (see Fig.~\ref{fig:boundaries}).

Throughout this paper, we will make the \emph{Linear Regions Assumption}: Each region represents a maximal connected component of input space on which the piecewise linear function $\N(\xx)$ is given by a single linear function. While this assumption has tacitly been made in the prior literature, it is noted in \citet{hanin2019deep} that there are cases where it does not hold. For example, if an entire layer of the network is zeroed out for some inputs, then $\N(\xx)$ may be given by the same linear function across several adjacent regions, despite these regions corresponding to different ReLU activation patterns.

\subsection{Main result}
\citet{hanin2019deep} show that for all but a measure-zero set of networks, $B_z$ is an $(\nin-1)$-dimensional piecewise linear surface in $\R^{\nin}$. We say that $B_z$ \emph{bends} at a point if $B_z$ is nonlinear at that point. As observed in \citet{hanin2019deep}, $B_z$ can bend only at points where it intersects boundaries $B_{z'}$ for $z'$ in an earlier layer of the network (see Fig.~\ref{fig:boundaries}, in which input dimension is 2 and the $B_z$ are simply bent lines). The following theorem (proven in Appendix \ref{app:thm_structure}) shows that the converse also holds.

\begin{thm}[Boundaries imply network structure] Let $\N$ be a fully connected ReLU network satisfying the Linear Regions Assumption. Then, the following statement holds except for $\N$ in a measure-zero set of networks: For every neuron $z$, the boundary $B_z$ bends exactly where it intersects a boundary $B_{z'}$ such that $z'$ is in an earlier layer than $z$.
\label{thm:structure}
\end{thm}

From this theorem, it follows that for any two intersecting boundaries $B_z$ and $B_{z'}$, one of the following must hold: $B_z$ bends at their intersection (in which case $z$ occurs in a deeper layer of the network), $B_{z'}$ bends (in which case $z'$ occurs in a deeper layer), or neither bends (in which case $z$ and $z'$ occur in the same layer). It is not possible for both $B_z$ and $B_{z'}$ to bend at their intersection -- unless that intersection is also contained in another boundary, which is vanishingly unlikely in general. Therefore, the architecture of the network can be determined by evaluating the boundaries $B_z$ and where they bend in relation to one another.

In fact, a much stronger statement holds; namely, the weights and biases of the network can also be determined from the boundaries:

\begin{thm}[Boundaries imply network weights] Let $\N$ be a fully connected ReLU network satisfying the Linear Regions Assumption. Suppose that for any two neurons $z$ and $z'$ that are connected in $\N$, the boundaries $B_z$ and $B_{z'}$ intersect. Then, given the set of boundaries between linear regions, it is possible to recover both the complete architecture of $\N$ and the weights and biases of every hidden layer, up to permutation and scaling, except for $\N$ in a measure-zero set of networks.
\label{thm:weights}
\end{thm}

This result follows from Theorem \ref{thm:correct}, which states that not only is it possible to recover the structure and weights of a network from the boundaries between linear regions, but the boundaries themselves can effectively be approximated by the algorithm we describe in \S\ref{sec:alg}, enabling the network to be reverse-engineered merely by querying it. We prove Theorem \ref{thm:correct} in Appendix \ref{app:thm_correct}.

As an intuition for why Theorem \ref{thm:weights} holds, observe first that for neurons $z$ in the first layer of the network, the boundaries $B_z$ are simply hyperplanes (see Fig.~\ref{fig:boundaries}). The equations of these hyperplanes reveal the weights from the input to the first layer (up to permutation, scaling, and sign). For each subsequent layer, the weight between neurons $z$ and $z'$ can be determined by calculating how $B_{z'}$ bends when it crosses $B_z$, as this dictates how much the input to $z$ changes when neuron $z'$ is zeroed out by its ReLU.

\section{Algorithm}
\label{sec:alg}

We now describe an algorithm to reverse-engineer a network $\N$ by approximating the boundaries between linear regions. Our algorithm assumes that the output of the network $\N(\xx)$ can be queried for different inputs $\xx$, but does not assume any \emph{a priori} knowledge of the linear regions or boundaries.

\label{sec:algorithm}
\begin{figure*}[ht]
\begin{center}
\includegraphics[scale=0.38]{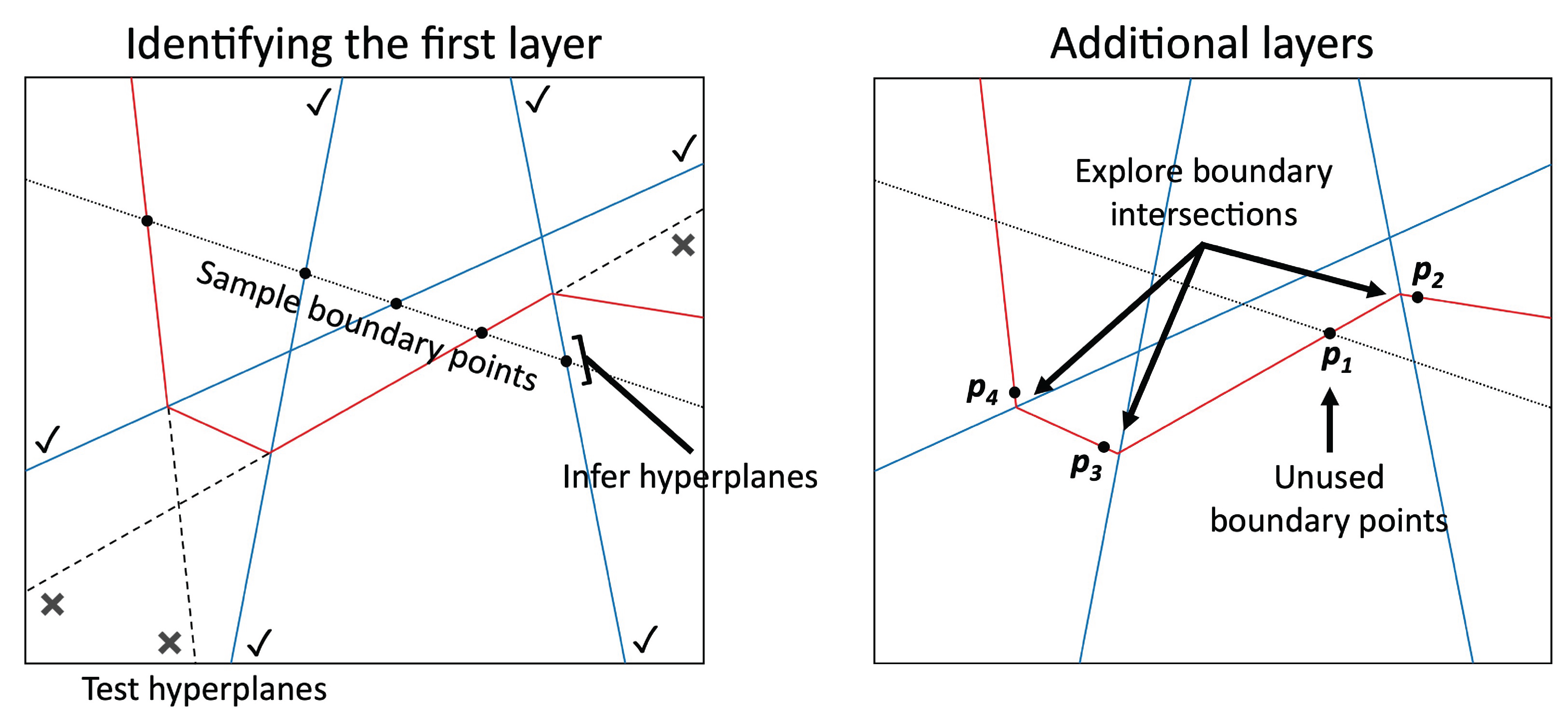}
\caption{Schematic of our algorithms for identifying the first layer and additional layers.}
\vspace{-0.2in}
\label{fig:algorithm}
\end{center}
\end{figure*}

\subsection{The first layer}
\label{subsec:first_layer}
We begin by identifying the first layer of the network $\N$, for which we must infer the number of neurons, the weight matrix $\WW^1$, and the bias vector $\bb^1$. As noted above, for each $z=z_i^1$ in the first layer, the boundary $B_z$ is a hyperplane with equation $\WW^1_{\cdot i}\xx + \bb^1_i = 0$. For each neuron $z$ in a later layer of the network, the boundary $B_z$ will, in general, bend and not be a (full) hyperplane (Theorem \ref{thm:structure}). We may therefore find the number of neurons in layer 1 by counting the hyperplanes contained in the network's boundary $B$, and we can infer weights and biases by determining the equations of these hyperplanes.
\begin{algorithm}[H]
\caption{The first layer}\label{fig:first_layer}
\begin{algorithmic}
\STATE Initialize $P_1=P_2=S_1=\{\}$
\FOR{$t=1,\ldots,L$}
	\STATE Sample line segment $\ell$
	\STATE $P_1\gets P_1\cup \texttt{PointsOnLine}(\ell)$
\ENDFOR
\FOR{$\pp \in P_1$}
	\STATE $H =\texttt{InferHyperplane}(\pp)$
	\IF{$\texttt{TestHyperplane}(H)$}
		\STATE $S_1\gets S_1 \cup \texttt{GetParams}(H)$
	\ELSE \STATE $P_2\gets P_2 \cup \{\pp\}$
	\ENDIF
\ENDFOR
\STATE $\text{\textbf{return} parameters }S_1,$\\
$\hskip .4 in\text{unused sample points }P_2$
\end{algorithmic}
\end{algorithm}
\vspace{-0.1in}

\textbf{Boundary points along a line.} Our algorithm is based upon the identification of points on the boundary $B$. One of our core algorithmic primitives is a subroutine \texttt{PointsOnLine} that takes as input a line segment $\ell \subset \R^\nin$ and approximates the set $\ell \cap B$ of boundary points along $\ell$. The algorithm proceeds by leveraging the fact that boundary points subdivide $\ell$ into regions within which $\N(\xx)$ is linear. We maintain a list of points in order along $\ell$ (initialized to the endpoints and midpoint of $\ell$) and iteratively perform the following operation: For each three consecutive points on our list, $\xx_1,\xx_2,\xx_3$, we determine if the vectors $(\N(\xx_2) - \N(\xx_1))/||\xx_2 - \xx_1||_2$ and $(\N(\xx_3) - \N(\xx_2))/||\xx_3 - \xx_2||_2$ are equal (to within computation error) -- if so, we remove the point $\xx_2$ from our list, otherwise we add the points $(\xx_1 + 2\xx_2)/3$ and $(\xx_3 + 2\xx_2)/3$ to our list.\footnote{These weighted averages speed up the search algorithm by biasing it towards the center of the segment, which is where we expect the most intersections given our choice of segments.} The points in the list converge by binary search to the set of discontinuities of the gradient $\nabla\N(\xx)$, which are our desired boundary points. Note that \texttt{PointsOnLine} is where we make use of our ability to query the network.

\textbf{Sampling boundary points.} In order to identify the boundaries $B_z$ for $z$ in layer 1, we begin by identifying a set of boundary points with at least one on each $B_z$. A randomly chosen line segment through input space will intersect some of the $B_z$ -- indeed, if it is long enough, it will intersect any fixed hyperplane with probability 1. We sample line segments $\ell$ in $\R^\nin$ and run \texttt{PointsOnLine} on each. Many sampling distributions are possible, but in our implementation we choose to sample segments of fixed (long) length, tangent at their midpoints to a sphere of fixed (large) radius.\footnote{The algorithm is not sensitive to the exact settings of these hyperparameters, provided they are relatively large.} This ensures that each of our sample lines remains far from the origin, where boundaries are in closer proximity and therefore more easily confused with one another (this will become useful in the next step). Let $P_1$ be the overall set of boundary points identified on our sample line segments.

\textbf{Inferring hyperplanes.} We now fit a hyperplane to each of the boundary points we have identified. For each $\pp\in P_1$, there is a neuron $z$ such that $\pp \in B_z$. The boundary $B_z$ is piecewise linear, with nonlinearities only along other boundaries, and with probability $1$, $\pp$ does not lie on a boundary besides $B_z$. Therefore, within a small enough neighborhood of $\pp$, $B_z$ is given by a hyperplane, which we call the \emph{local hyperplane} at $\pp$. If $z$ is in layer 1, then $B_z$ equals the local hyperplane. The subroutine \texttt{InferHyperplane} takes as input a point $\pp$ on a boundary $B_z$ and approximates the local hyperplane within which $\pp$ lies. This algorithm proceeds by sampling many small line segments around $\pp$, running \texttt{PointsOnLine} to find their points of intersection with $B_z$, and performing linear regression to find the equation of the hyperplane containing these points.

\textbf{Testing hyperplanes.} Not all of the hyperplanes we have identified are actually boundaries for neurons in layer 1, so we need to test which hyperplanes are contained in $B$ in their entirety, and which are the local hyperplanes of boundaries that bend. The subroutine \texttt{TestHyperplane} takes as input a point $\pp$ and a hyperplane $H$ containing that point, and determines whether the entire hyperplane $H$ is contained in the boundary $B$ of the network. This algorithm proceeds by sampling points within $H$ that lie far from $\pp$ and applying \texttt{PointsOnLine} to a short line segment around each such point to check whether these points all lie on $B$. Applying \texttt{TestHyperplane} to those hyperplanes inferred in the preceding step allows us to determine those $B_z$ for which $z$ is in layer 1.

\textbf{From hyperplanes to parameters.} Finally, we identify the first layer of $\N$ from the equations of hyperplanes contained in $B$. The number of neurons in layer 1 is given simply by the number of distinct $B_z$ that are hyperplanes.  As we have observed, for $z=z_i^1$ in layer 1, the hyperplane $B_z$ is given by $\WW^1_{\cdot i}\xx + \bb^1_i = 0$. We can thus determine $\WW^1_{\cdot i}$ and $\bb^1_i$ up to multiplication by a constant. However, we have already observed that scaling $\WW^1_{\cdot i}$ and $\bb^1_i$ by a positive constant (while inversely scaling $\WW^2_{i\cdot}$) is a network isomorphism (\S\ref{subsec:isomorphisms}). Therefore, we need only determine the true sign of the multiplicative constant, corresponding to determining which side of the hyperplane is zeroed out by the ReLU. This determination of sign will be performed in \S\ref{subsec:other_layers}.

\subsection{Additional layers}
\label{subsec:other_layers}
We now assume that the weights $\WW^1, \ldots, \WW^{k-1}$ and biases $\bb^1, \ldots, \bb^{k-1}$ have already been determined within the network $\N$, with the exception of the sign choice for weights and biases at each neuron in layer $k-1$. We now show how it is possible to determine the weights $\WW^k$ and biases $\bb^k$, along with the correct signs for $\WW^{k-1}$ and $\bb^{k-1}$.
\begin{algorithm}[H]
\caption{Additional layers}\label{fig:additional_layers}
\begin{algorithmic}
\STATE Input $P_k$ and $S_1,\ldots,S_{k-1}$
\STATE Initialize $S_k=\{\}$
\FOR{$\pp_1 \in P_{k-1}$ on boundary $B_z$}
	\STATE Initialize $A_z = \{\pp_1\}$, $L_z = \Hh_z=\{\}$
	\WHILE{$L_z \not\supseteq \text{Layer }k-1$}
	\STATE Pick $\pp_i\in A$ and $\vv$
	\STATE $\pp',B_{z'}=\texttt{ClosestBoundary}(\pp_i,\vv)$
	\IF{$\pp'$ on boundary}
	\STATE $A_z\gets A_z\cup \{\pp' + {\bf\epsilon}\}$
	\STATE $L_z\gets L_z\cup \{z'\}$
	\STATE $\Hh_z\gets \Hh_z\cup \{\texttt{InferHyperplane}(\pp_i)\}$
	\ELSE \STATE $P_k\gets P_k\cup \{\pp_1\}$; \textbf{break}
	\ENDIF
	\ENDWHILE
	\IF{$L_z\supseteq \text{Layer }k-1$}
		\STATE $S_k\gets \texttt{GetParams}(T_z)$
	\ENDIF
\ENDFOR
\STATE $\text{\textbf{return} parameters }S_k,\text{unused sample points }P_{k+1}$
\end{algorithmic}
\end{algorithm}
\vspace{-0.1in}
\textbf{Closest boundary along a line.} In this part of our algorithm, we will need the ability to move along a boundary to its intersection with another boundary. For this purpose, the subroutine \texttt{ClosestBoundary} will be useful. It takes as input a point $\pp$, a vector $\vv$ and the network parameters as determined up to layer $k-1$, and outputs the smallest $c>0$ such that $\qq = \pp + c\vv$ lies on $B_z$ for some $z$ in layer at most $k-1$. In order to compute $c$, we consider the region $R$ within which $\pp$ lies, which is associated with a certain pattern of active and inactive ReLUs. For each boundary $B_z$, we calculate the hyperplane equation which would define $B_z$ were it to intersect $R$, due to the fixed activation pattern within $R$, and we calculate the distance from $\pp$ to this hyperplane. While not every boundary $B_z$ intersects $R$, the closest boundary does, allowing us to find the desired $c$.

\textbf{Unused boundary points.} In order to identify the boundaries $B_z$ for $z$ in layer $k$, we wish to identify a set of boundary points with at least one on each such boundary. However, in previous steps of our algorithm, a set $P_{k-1}$ of boundary points was created, of which some were used in ascertaining the parameters of earlier layers. We now consider the subset $P_k \subset P_{k-1}$ of points that were not found to belong to $B_z$, for $z$ in layers $1$ through $k-1$. These points have already had their local hyperplanes determined.

\textbf{Exploring boundary intersections.} Consider a point $\pp_1\in P_k$ such that $\pp_1 \in B_z$. Note that $B_z$ will, in general, have nonlinearities where it intersects $B_{z'}$ such that $z'$ lies in an earlier layer than $z$. We explore these intersections, and in particular attempt to find a point of $B_z \cap B_{z'}$ for every $z'$ in layer $k-1$. Given the local hyperplane $H$ at $\pp_1$, we pick a direction $\vv$ along $H$ and apply \texttt{ClosestBoundary} to calculate the closest point of intersection $\pp'$ with $B_{z'}$ for all $z'$ already identified in the network. (We discuss below how to pick $\vv$.) Note that if $z$ is in layer $k$, then $\pp'$ must be on $B_z$ as well as $B_{z'}$, while if $z$ is in a later layer of the network, then there may exist unidentified neurons in layers below $z$ and therefore $B_z$ may bend before meeting $B_{z'}$. We check if $\pp'$ lies on $B_z$ by applying \texttt{PointsOnLine}, and if so apply \texttt{InferHyperplane} to calculate the local hyperplane of $B_z$ on the other side of $B_{z'}$ from $\pp_1$. We select a representative point $\pp_2$ on this local hyperplane.  We repeat the process of exploration from the points $\pp_1, \pp_2, \ldots$ until one of the following occurs: (i) a point of $B_z \cap B_{z'}$ has been identified for every $z'$ in layer $k-1$ (this may be impossible; see \S\ref{subsec:boundaries}), (ii) $z$ is determined to be in a layer deeper than $k$ (as a result of $\pp'$ not lying on $B_z$), or (iii) a maximum number of iterations has been reached.

\textbf{How to explore.} An important step in our algorithm is exploring points of $B_z$ that lie on other boundaries. Given a set of points $A_z = \{\pp_1, \pp_2, \ldots, \pp_m\}$ on $B_z$, there are several methods for picking a point $\pp_i$ and direction $\vv$ along the local hyperplane at $\pp_i$ to apply \texttt{ClosestBoundary}. One approach is to pick a random point $\pp_i$ from those already identified and a random direction $\vv$; this has the advantage of simplicity. However, it is somewhat faster to consider for which $z'$ the intersection $B_z\cap B_{z'}$ has not yet been identified and attempt specifically to find points on these intersections. One approach for this is to pick a missing $z'$ and identify for which $\pp_i$ the boundary $B_{z'}$ lies on the boundary of the region containing $\pp_i$ and solve a linear program to find $\vv$. Another approach is to pick a missing $z'$ and a point $\pp_i$, calculate the hyperplane $H$ which would describe $B_{z'}$ under the activation pattern of $\pp_i$, and choose $\vv$ along the local hyperplane to $\pp_i$ such that the distance to $H$ is minimized. This is the approach which we take in our implementation, though more sophisticated approaches may exist and present an interesting avenue for further work.

\textbf{From boundaries to parameters.} We now identify layer $k$ of $\N$, along with the sign of the parameters of layer $k-1$, by measuring the extent to which boundaries bend at their intersection. We are also able to identify the correct signs at layer $k-1$ by solving an overconstrained system of constraints capturing the influence of neurons in layer $k-1$ on different regions of input space. Theorem \ref{thm:correct} formalizes the inductive step that allows us to go from what we know at layer $k-1$ (weights and biases, up to scaling and sign) to the equivalent set of information for layer $k$.

\section{Discussion}
\subsection{Correctness and sample complexity}
\label{subsec:complexity}
The following theorem (proven in Appendix \ref{app:thm_correct}) establishes the validity of our algorithm above.

\begin{thm}[Algorithm proof of correctness] The above algorithm successfully recovers, up to permutation and scaling, the structure and weights of any deep ReLU network for which the assumptions of Theorem \ref{thm:weights} hold. No prior knowledge of the weights, structure, or linear region boundaries is assumed, merely the ability to query the network.

Specifically, the following holds true for fully connected ReLU networks $\N$ satisfying the Linear Region Assumption (\S\ref{subsec:regions}), excluding a set of networks with measure zero: Suppose that the weights and biases of $\N$ are known up through layer $k-1$, with the exception that for each neuron in layer $k-1$, the sign of the incoming weights and the bias is unknown. Suppose also that for each $z$ in layer $k$, there exists an ordered set of points $A_z= \{\pp_1, \pp_2, \ldots, \pp_m\}$ such that: (i) Each point lies on the boundary of $B_z$, and in (the interior of) a distinct region with respect to the earlier-layer boundaries already known; (ii) each point (except for $\pp_1$) has a precursor in an adjacent region; (iii) for each such pair of points, the local hyperplanes of $B_z$ are known, as is the boundary $B_{z'}$ dividing them ($z'$ in an earlier layer); (iv) the set of such $z'$ includes all of layer $k-1$.

Then, it is possible to recover the weights and biases for layer $k$, with the exception that for each neuron, the sign of the incoming weights and the bias is unknown. It is also possible to recover the sign for every neuron in layer $k-1$.
\label{thm:correct}
\end{thm}

Note that even when assumption (iv) of the Theorem is violated, the algorithm recovers the weights corresponding to whichever boundaries are successfully crossed, as we verify empirically in \S\ref{sec:experiments}.

We expect the number of queries necessary to obtain weights and biases (up to sign) for the first layer should grow as $O(\nin (\sum_i n_i)\log n_1)$, which for constant-width networks is only slightly above the number of parameters being inferred. Namely, if the biases in the network are bounded above, then each sufficiently long line has constant probability of hitting a given hyperplane, suggesting $\log n_1$ lines are required according to a coupon collector-style argument. \citet{hanin2019complexity} prove that under natural assumptions, the number of boundary points intersecting a given line through input space grows linearly in the total number of neurons in the network. Finally, each boundary point on a line requires $O(\nin)$ queries in order to fit a hyperplane.

For deeper layers, the number of queries depends on the approach taken to explore the intersections between boundaries. It is possible, depending on the arrangement of intersections, for the number of queries to grow linearly in the number of parameters, as each weight can be inferred by examining a single intersection between boundaries.

\subsection{Assumptions}
\label{subsec:boundaries}
It is possible that for some neurons $z$ and $z'$ in consecutive layers, there is no point of intersection between the boundaries $B_z$ and $B_{z'}$ (or that this intersection is very small), making it impossible to infer the weight between $z$ and $z'$ by our algorithm. Some such cases represent an ambiguity in the underlying network -- an additional isomorphism to those described in \S\ref{subsec:isomorphisms}. Namely, $B_z\cap B_{z'}$ is empty if one of the following cases holds: (1) whenever $z$ is active, $z'$ is inactive; (2) whenever $z$ is active, $z'$ is active; (3) whenever $z$ is inactive, $z'$ is inactive; or (4) whenever $z$ is inactive, $z'$ is active.  In case 1, observe that a slight perturbation to the weight $w$ between $z$ and $z'$ has no effect upon the network's output; thus $w$ is not uniquely determined. Cases 2-4 present a more complicated picture; depending on the network, there may or may not be additional isomorphisms.

For simplicity in our algorithm, we have not considered the relatively rare cases where boundaries $B_z$ are disconnected or bounded. If $B_z$ is disconnected, then it may not be possible to find a connected path along it that intersects all boundaries arising from the preceding layer. In this case, it is simple to infer that two independently identified pieces of the boundary belong to the same neuron to infer the full weight vector.  Next, if $B_z$ is bounded for some $z$, then it is a closed $(d-1)$-dimensional surface within $d$-dimensional input space\footnote{For 2D input, such $B_z$ must be topological circles, but for higher dimensions, it is conceivable for them to be more complicated surfaces, such as toroidal polyhedra.}. While our algorithm requires no modification in this case, bounded $B_z$ may be more difficult to find by intersection with randomly chosen lines, and a more principled sampling method may be helpful.

\subsection{Other architectures} While we have expressed our algorithm in terms of multi-layer perceptrons with ReLU activation, it also extends to various other architectures of neural network. Other piecewise linear activation functions admit similar algorithms. For a network with convolutional layers, it is possible to use the same approach to infer the weights between neurons, with two caveats: (i) As we have stated it, the algorithm does not account for weight-sharing -- the number of ``neurons'' in each layer is thus dependent on the input size, and is very large for reasonably sized images. (ii) Pooling layers \emph{do} affect the partition into activation regions, and indeed introduce new discontinuities into the gradient; our algorithm therefore does not apply.

For skip connections as in ResNets \citep{he2016deep}, our algorithm should hold with slight modifications, which we outline here. As in a multi-layer perceptron, each boundary component $B_z$ is a bent hyperplane that bends when it intersects $B_{z'}$ for $z'$ at an earlier layer than $z$. However, potential weights must in this case be considered between neurons in any two different layers. Deriving the weights for such skip connections is somewhat more complex than for multi-layer perceptrons, as the ``bend'' is influenced not merely by the skip connection but by the weights along all other paths between the two neurons through the network. Thus, it is necessary to “move backward” through the network -- for a neuron in layer $k$, one must first derive the weights of connections arising from the preceding layer $k-1$, then from $k-2$, and so on.

\section{Experiments}
\label{sec:experiments}
We demonstrate the success of our algorithm on both untrained and trained networks. In keeping with literature on ReLU network initialization \citep{he2015delving,hanin2018start}, networks were initialized using i.i.d.~normal weights with variance $2/\text{fan-in}$ and i.i.d.~normal biases with unit variance. Networks were trained on either the MNIST dataset ($\nin = 784,\nout = 10$) or a memorization task of 1000 ``datapoints'' ($\nin = 10,\nout = 2$) with coordinates drawn i.i.d.~from a unit Gaussian and given arbitrary binary labels. Training was performed using the Adam optimizer \citep{kingma2014adam} and a cross-entropy loss applied to the softmax of the final layer, over 20 epochs for MNIST and 1000 epochs for the memorization task. The trained networks (when sufficiently large) were able to attain near-perfect accuracy.
\begin{figure*}[ht]
\begin{center}
\includegraphics[scale=0.24]{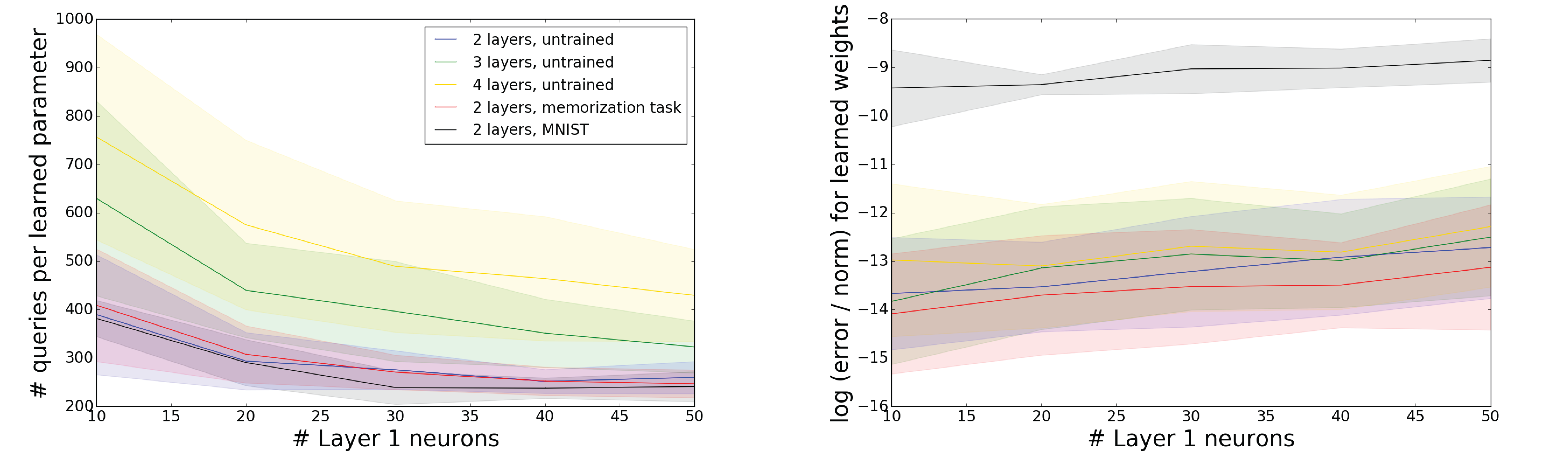}
\vspace{-0.2in}
\caption{Results of our first-layer algorithm, applied to networks with two or more hidden layers as the width of the first layer varies. All other layers have fixed width 10. Untrained networks have input and output dimension 10, those trained on the memorization task have input dimension 10 and output dimension 2, and those trained on MNIST have input dimension 784 and output dimension 10. Left: The number of queries issued by our algorithm per parameter identified in the network $\N$; the algorithm is terminated once the desired number of neurons have been identified. Right: Log normalized error $\log(||\hat{\WW}^1 - \WW^1||_2 / ||\hat{\WW^1}||_2)$ for $\hat{\WW}^1$ the approximated weights. Weight vectors were scaled to unit norm to account for isomorphism (see \S\ref{subsec:isomorphisms}).  Curves are averaged over 5 runs in the case of MNIST and 40 runs otherwise, with standard deviations between runs shown as shaded regions.}
\vspace{-0.1in}
\label{fig:layer1}
\end{center}
\end{figure*}

\begin{figure*}[ht]
\begin{center}
\includegraphics[scale=0.24]{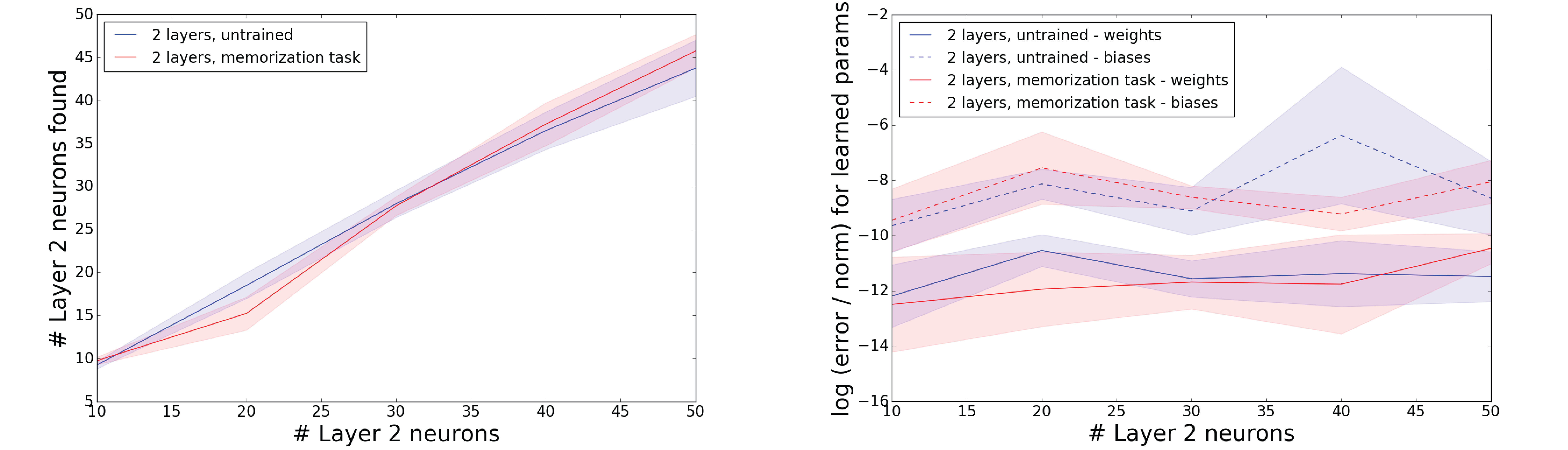}
\vspace{-0.2in}
\caption{Results of our algorithm for additional layers, applied to networks with two layers, the first layer of width 10, as the width of the second layer varies. Left: Number of estimated layer-2 neurons. Right: Log normalized error between estimated and corresponding true neurons for approximated layer-2 weights and biases. Curves are averaged over 4 runs, with standard deviations shown as shaded regions.}
\vspace{-0.2in}
\label{fig:layer2}
\end{center}
\end{figure*}

We observe that both the first-layer algorithm and additional-layer algorithm identified weights and biases to within extremely high accuracy (see Figs.~\ref{fig:layer1} and \ref{fig:layer2}). In order to compare estimated and true parameters, we scaled weights at each neuron to unit norm and accounted for possible permutations of the neurons within a layer. (As described in \S\ref{subsec:isomorphisms}, these transformations do not change the function computed by the network and represent unavoidable isomorphisms in recovering the network.) Figures show the log normalized error $\log(||\hat{\WW}^k - \WW^k||_2 / ||\hat{\WW^k}||_2)$ between true weights $\WW^k$ and approximated weights $\hat{\WW}^k$ and likewise the log normalized error $\log(||\hat{\bb}^k - \bb^k||_2 / ||\hat{\bb^k}||_2)$ between true biases $\bb^k$ and approximated weights $\hat{\bb}^k$.

In keeping with our analysis in \S\ref{subsec:complexity}, the number of queries needed to recover the first layer of a network scales linearly with the number of neurons in that layer (Fig.~\ref{fig:layer1}). In the second layer, a small fraction of weights are sometimes not identified (see \S\ref{subsec:boundaries} for a discussion of reasons why such behavior is inevitable); even in such cases, the algorithm is able to correctly predict the remaining parameters (Fig.~\ref{fig:layer2}).

\section{Conclusion}
In this work, we have proven that it is possible to recover the architecture, weights, and biases of deep ReLU networks from the boundaries between linear regions defined by the network. In many cases, we show that it is possible to reconstruct these boundaries and thus the network itself merely by querying the network's output on certain inputs. Networks can, we observe, often be reconstructed unambiguously up to permutation of neurons within each layer and scaling of weights and biases at individual neurons. In some cases, there are additional isomorphisms, in which case our algorithm is able to reconstruct a significant fraction of the parameters of the network.

Our approach works for a wide variety of networks, though not all. It is limited to ReLU or otherwise piecewise linear activation functions, though we believe it possible that a continuous version of this method could potentially be developed in future work for use with sigmoidal activation. If used with convolutional layers, our method does not account for the symmetries of the network and therefore scales with the size of the input as well as the number of features, resulting in high computation. Finally, the method is not robust to defenses such as adding noise to the outputs of the network, and therefore can be thwarted by a network designer that seeks to hide their weights/architecture.

We believe that the methods we have introduced here will lead to considerable advances in reverse-engineering neural networks, both in the context of deep learning and, more speculatively, in neuroscience. While the implementation we have demonstrated here is effective in small instances, we anticipate future work that optimizes these methods for efficient use with different architectures and at scale.

\bibliography{references}
\bibliographystyle{icml2020}
\vfill
\pagebreak
\appendix

\section{Isomorphism under scaling}
\label{app:lem}

\begin{lem}Given a fully connected ReLU network $\N$, the network $s_{z,c}(\N)$ is isomorphic to $\N$ for every neuron $z$ and constant $c>0$.
\label{lem:scaling}
\end{lem}
Suppose that $z=z_i^k$ is the $i$th neuron in layer $k$.  Then, for each neuron $z_j^{k+1}$ in layer $k+1$ of the network $\N$, we have:
\begin{align}
z_j^{k+1}(\xx) &= \sum_{i=1}^{n_k} \WW^k_{ij}\ReLU(z_i^k(\xx) + \bb^k_i)\nonumber \\
&= \sum_{i=1}^{n_k} \WW^k_{ij}\ReLU\Biggl(\Biggl(\sum_{h=1}^{n_{k-1}} \WW^{k-1}_{hi}\ReLU(z_h^{k-1}(\xx) + \bb^{k-1}_h)\Biggr) + \bb^k_i\Biggr).\label{eqn:net1}
\end{align}

By comparison, in network $s_{z,c}(\N)$, we have:
\begin{align}
z_j^{k+1}(\xx) &= \sum_{i=1}^{n_k} \frac{1}{c}\WW^k_{ij}\ReLU(z_i^k(\xx) + c\bb^k_i)\nonumber \\
&= \sum_{i=1}^{n_k} \frac{1}{c}\WW^k_{ij}\ReLU\Biggl(\Biggl(\sum_{h=1}^{n_{k-1}} c\WW^{k-1}_{hi}\ReLU(z_h^{k-1}(\xx) + \bb^{k-1}_h)\Biggr) + c\bb^k_i\Biggr)\nonumber \\
&= \sum_{i=1}^{n_k} \WW^k_{ij}\ReLU\Biggl(\Biggl(\sum_{h=1}^{n_{k-1}} \WW^{k-1}_{hi}\ReLU(z_h^{k-1}(\xx) + \bb^{k-1}_h)\Biggr) + \bb^k_i\Biggr),\label{eqn:net2}
\end{align}
where we used the property that $\ReLU(cx) = c\ReLU(x)$ for any $c>0$.

As expressions (\ref{eqn:net1}) and (\ref{eqn:net2}) are equal, we conclude that $s_{z,c}(\N)$ is isomorphic to $\N$.

\section{Proof of Theorem \ref{thm:structure}}
\label{app:thm_structure}
It is observed in \citet{hanin2019deep} that $B_z$ cannot bend except at points of intersection with $B_{z'}$ for $z'$ in an earlier layer than $z$. We now prove the converse. Suppose that neurons $z,z'$ are such that $z'$ lies in an earlier layer than $z$. Consider a point $\pp$ of intersection between $B_z$ and $B_{z'}$, and suppose that $\pp_1$ and $\pp_2$ are in an arbitrarily small neighborhood of $\pp$, lying on opposite sides of $B_{z'}$. It suffices to prove that $\nabla z(\pp_1)\ne \nabla z(\pp_2)$, and therefore that $B_z$ bends as it intersects $B_{z'}$.

By the Linear Regions Assumption, $\N(\xx)$ computes different functions on the two sides of $B_{z'}$. Since $\N(\xx)$ is continuous, $\nabla \N(\xx)$ must differ on the two sides of $B_{z'}$; that is, $\nabla \N(\pp_1)\ne \nabla\N(\pp_2)$. Suppose that $z=z_j^k$ lies in layer $k$; then there exists some neuron $z_\ell^k$ in layer $k$ such that $\nabla z_\ell^k(\pp_1)\ne \nabla z_\ell^k(\pp_2)$. If $j=\ell$, we are done. Otherwise, observe that
$$z_\ell^k(\xx) = \sum_{h=1}^{n_{k-1}} \WW^k_{h\ell}\ReLU(z_h^{k-1}(\xx) + \bb^{k-1}_h).$$

Consider the $\nin \times n_{k-1}$ matrix $M(\xx)$ with columns $\nabla\ReLU(z_h^{k-1}(\xx) + \bb^{k-1}_h)$ indexed by $h$. As $\nabla z_\ell^k(\xx)$ is a linear combination of these columns, we conclude that $M(\pp_1)\ne M(\pp_2)$. Note that $\nabla z(\xx) = \nabla z_j^k(\xx)$ is a linear combination of the columns of $M(\xx)$ with coefficients $\WW^k_{hj}$. Since $M(\pp_1)\ne M(\pp_2)$, we conclude that with probability 1 over the choice of $\WW^k_{hj}$, we must have $\nabla z(\pp_1)\ne \nabla z(\pp_2)$, as desired.

\section{Proof of Theorem \ref{thm:correct}}
\label{app:thm_correct}
In this proof, we will show how the information we are given by the assumptions of the theorem is enough to recover the weights and biases for each neuron $z$ in layer $k$. We will proceed for each $z$ individually, progressively learning weights between $z$ and each of the neurons in the preceding layer (though for skip connections this procedure could also easily be generalized to learn weights from $z$ to earlier layers).

For each of the points $\pp_i \in A_z$, suppose that $H_i$ is the local hyperplane associated with $\pp_i$ on boundary $B_z$. The gradient $\nabla z(\pp_i)$ at $\pp_i$ is orthogonal to $H_i$, and we thus already know the direction of the gradient, but its magnitude is unknown to us. We will proceed in order through the points $\pp_1,\pp_2,\ldots,\pp_m$, with the goal of identifying $\nabla z(\pp_i)$ for each $\pp_i$, up to a single scaling factor, as this computation will end up giving us the incoming weights for $z$.

We begin with $\pp_1$ by assigning $\nabla z(\pp_1)$ arbitrarily to either one of the two unit vectors orthogonal to $H_i$. Due to scaling invariance (Lemma \ref{lem:scaling}), the weights of $\N$ can be rescaled without changing the function so that $\nabla z(\pp_i)$ is multiplied by any positive constant. Therefore, our arbitrary choice can be wrong at most in its sign, and we need not determine the sign at this stage.  Now, suppose towards induction that we have identified $\nabla z(\pp_i)$ (up to sign) for $i=1,\ldots,s-1$. We wish to identify $\nabla z(\pp_s)$.

By assumption (ii), there exists a precursor $\pp_r$ to $\pp_s$ such that $H_r$ and $H_s$ intersect on a boundary $B_{z'}$. Let $\vv_r = t_z\nabla z(\pp_r)$ be our estimate of $\nabla z(\pp_r)$, for unknown sign $t_z \in \{+1, -1\}$. Let $\vv_s$ be a unit normal vector to $H_s$, so that $\vv_s = ct_z\nabla z(\pp_s)$ for some unknown constant $c$.  We pick the sign of $\vv_s$ so that it has the same orientation as $\vv_r$ with respect to the surface $B_z$, and thus $c>0$.  Finally, let $\vv = t_{z'}\nabla z'(\pp_r) = t_{z'}\nabla z'(\pp_s)$ be our estimate of the gradient of $z'$; where $t_{z'} \in \{+1, -1\}$ is also an unknown sign (recall that since $z'$ is in layer $k-1$ we know its gradient up to sign).  We will use $\vv$ and $\vv_r$ to identify $\vv_s$.

Suppose that $z=z_j^k$ is the $j$th neuron in layer $k$ and that $z'=z_h^{k-1}$ is the $h$th neuron in layer $k-1$. Recall that
\begin{equation}
z(\xx) = z_j^k(\xx) = \sum_{i=1}^{n_{k-1}} \WW^k_{ij}\ReLU(z_i^{k-1}(\xx) + \bb^{k-1}_i).\label{eqn:layer}
\end{equation}
As $B_{z'}$ is the boundary between inputs for which $z'=z_h^{k-1}$ is active and inactive, $\ReLU(z_h^{k-1}(\xx) + \bb^{k-1}_h)$ must equal zero either (Case 1) on $H_r$ or (Case 2) on $H_s$.

In Case 1, we have
$$
\nabla z(\pp_s) - \nabla z(\pp_r) = \WW^k_{hj}\nabla z'(\pp_r),
$$
or equivalently:
$$
ct_z\vv_s - t_z\vv_r = \WW^k_{hj} t_{z'} \vv,
$$
which gives us the equation:
$$
c\vv_s - \vv_r = \WW^k_{hj} t_zt_{z'} \vv.
$$
Since we know the vectors $\vv_s, \vv_r, \vv$, we are able to deduce the constant $c$.

A similar equation arises in Case 2:
$$
\vv_r - c\vv_s = \WW^k_{hj} t_zt_{z'} \vv,
$$
giving rise to the same value of $c$. We thus may complete our induction. In the process, observe that we have calculated a constant $\WW^k_{hj} t_zt_{z'}t'$, where the sign $t'$ is $+1$ in Case 1 and $-1$ in Case 2.  Note that $t_{z'}t'$ can be calculated based on whether $\vv$ points towards $\pp_r$ or $\pp_s$.  Therefore, we have obtained $\WW^k_{hj}t_z$, which is exactly the weight (up to $z$-dependent sign) that we wished to find.  Once we have all weights incoming to $z$ (up to sign), it is simple to identify the bias for this neuron (up to sign) by calculating the equation of any known local hyperplane for $B_z$ and using the known weights and biases from earlier layers.

To complete the proof, we must now also calculate the correct signs $t_{z'}$ of the neurons in layer $k-1$. Pick some $z=z_j^k$ in layer $k$ and observe that for all points $\pp_s\in A_z$ there corresponds an equation, obtained by taking gradients in equation (\ref{eqn:layer}):
$$
\nabla z_j^k(\pp_s) = \sum_{i=1}^{n_{k-1}} \WW^k_{ij}\One_{i,s}\nabla z_i^{k-1}(\pp_s),
$$
where $\One_{i,s}$ equals $1$ if $\pp_s$ is on the active side of $B_{z_i^{k-1}}$.  We can substitute in our (sign-unknown) values for these various quantities:
$$
t_z\vv_s = \sum_{i=1}^{n_{k-1}} \WW^k_{ij}\One_{i,s}t_{z_i^{k-1}}\vv_i.
$$

Now, we may estimate $\One_{i,s}$ by a function $\One'_{i,s}$ that is 1 if $\pp_s$ and $\vv_i$ are on the same side of $B_{z_i^{k-1}}$.  This estimate will be wrong exactly when $t_{z_i^{k-1}}=-1$.  Thus, $\One_{i,s} = (1 + t_{z_i^{k-1}} \One'_{i,s}) / 2$, giving us the equation:
\begin{align*}
t_z\vv_s &= \sum_{i=1}^{n_{k-1}} \WW^k_{ij}\frac{1 + t_{z_i^{k-1}} \One'_{i,s}}{2}t_{z_i^{k-1}}\vv_i\\
&=\frac{1}{2}\sum_{i=1}^{n_{k-1}} \WW^k_{ij}(t_{z_i^{k-1}} + \One'_{i,s})\vv_i
\end{align*}
All the terms of this equation are known, with the exception of $t_z$ and the $n_{k-1}$ variables $t_{z_i^{k-1}}$ -- giving us a linear system in $n_{k-1} + 1$ variables. For a given $z_j^k$, there are $n_{k-1}$ different $\pp_s$ representing the intersections with $B_{z'}$ for each $z'$ in layer $k-1$; choosing these $\pp_s$ should in general give linearly independent constraints.  Moreover, the equation is in fact a vector equality with dimension $\nin$; hence, it is a highly overconstrained system, enabling us to identify the signs $t_{z_i^{k-1}}$ for each $z_i^{k-1}$.  This completes the proof of the theorem.
\end{document}